\documentclass[10pt, a4paper]{article}

\usepackage{lrec-coling2024} 
\usepackage{algorithm}
\usepackage{algpseudocode}
\usepackage{amsmath}
\usepackage{multirow}
\usepackage{times}
\usepackage{latexsym}
\usepackage{kotex}
\usepackage{graphicx}
\usepackage{bbding}
\usepackage{pifont}
\usepackage{wasysym}
\usepackage{amssymb}
\usepackage{enumitem}
\usepackage[T1]{fontenc}
\usepackage{color, colortbl}
\usepackage{tabularx}
\usepackage{makecell}
\usepackage{xcolor}

\usepackage[utf8]{inputenc}

\usepackage{microtype}

\usepackage{inconsolata}

\usepackage{graphicx}
\usepackage{booktabs}
\usepackage{times}
\usepackage{latexsym}
\usepackage{kotex}
\usepackage{graphicx}
\usepackage{bbding}
\usepackage{pifont}
\usepackage{wasysym}
\usepackage{amssymb}
\usepackage{enumitem}
\usepackage{algorithm}
\usepackage{algpseudocode}
\usepackage{amsmath}
\usepackage{multirow}
\usepackage{subfig}
\usepackage{amsmath,amssymb,amsthm}
\usepackage{hyperref}
\hypersetup{
    colorlinks=true,
    linkcolor=blue,
    filecolor=magenta,      
    urlcolor=blue,
}

\usepackage{color}

\usepackage[T1]{fontenc}
\usepackage{makecell}

\usepackage[utf8]{inputenc}

\usepackage{microtype}

\usepackage{inconsolata}
\usepackage{graphicx}

 \usepackage{ulem}

\title{SentiCSE: A Sentiment-aware Contrastive Sentence Embedding Framework with Sentiment-guided Textual Similarity}

\name{Jaemin Kim\thanks{*Co-first authors.}$^{1,2*}$, Yohan Na$^{1*}$, Kangmin Kim\thanks{$^{2,3}$This work was done while the authors were at Hanyang University. Currently, Jaemin Kim and Kangmin Kim are working as NLP research engineers at LG Electronics and BHSN respectively.}$^{1,3}$, Sang Rak Lee$^1$, Dong-Kyu Chae\thanks{$^\dagger$Corresponding author.}$^{1\dagger}$} 

\address{$^1$Hanyang University, $^2$LG Electronics, $^3$BHSN \\
         \{jaemink, nayohan, kevin7133, sangrak, dongkyu\}@hanyang.ac.kr\\\         
         }

\abstract{
Recently, sentiment-aware pre-trained language models (PLMs) demonstrate impressive results in downstream sentiment analysis tasks.
However, they neglect to evaluate the quality of their constructed sentiment representations; 
they just focus on improving the fine-tuning performance, which overshadows the representation quality.
We argue that without guaranteeing the representation quality, their downstream performance can be highly dependent on the supervision of the fine-tuning data
rather than representation quality.
 This problem would make them difficult to foray into other sentiment-related domains, especially where labeled data is scarce. We first propose \textit{Sentiment-guided Textual Similarity} (SgTS), a novel metric for evaluating the quality of sentiment representations, which is designed based on the degree of equivalence in sentiment polarity between two sentences. We then propose SentiCSE, a novel Sentiment-aware Contrastive Sentence Embedding framework for constructing sentiment representations via combined word-level and sentence-level objectives, whose quality is guaranteed by SgTS. Qualitative and quantitative comparison with the previous sentiment-aware PLMs shows the superiority of our work. Our code is available at: \href{https://github.com/nayohan/SentiCSE}{https://github.com/nayohan/SentiCSE}
 \\ \newline  }

\begin{document}

\maketitleabstract

\section{Introduction}

\textit{Sentiment analysis} has been attracting great attention from researchers in not only the field of natural language processing (NLP) but also various other domains such as recommender systems, management information systems, and social media analysis. Sentiment analysis can play a critical role in the decision-making process by obtaining public opinion about specific products, services, or social issues~\citep{zhang2018deep,qi2015sentiment}.
In particular, more and more people are actively expressing their opinions online these days, which makes sentiment analysis even more important~\citep{yadav2020sentiment}.

\begin{figure}[t]
    \centerline{\includegraphics[width= 0.45\textwidth]{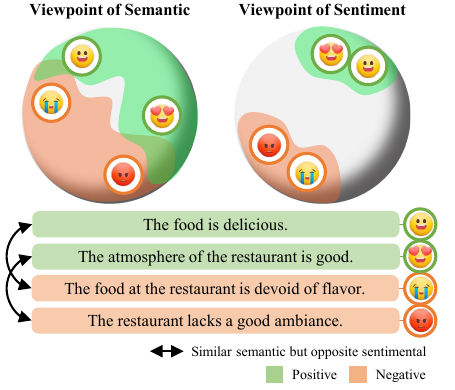}}
    \vspace{-2mm}
    \caption{The difference between sentence embedding from a semantic perspective and a sentiment perspective, which shows the necessity of focusing on embedding methods from a sentiment perspective for sentiment analysis, instead of the traditional semantic perspective.}
   \label{fig:squirrel}
   \vspace{-5mm}
   \label{fig:fig1}
\end{figure}

Recently, pre-trained language models (PLMs) such as BERT \citep{devlin-etal-2019-bert} and GPT \citep{radford2018improving} have accomplished remarkable performance in various NLP tasks by capturing semantic information and constructing contextualized representations. Under this guiding trend, a number of researches on sentiment analysis employed the PLMs as a backbone and made the sentiment-related applications as a downstream task. 
However, directly employing the PLMs has a limitation in sentiment analysis because there is a significant difference between \textit{textual similarity in terms of semantics} and \textit{textual similarity from the viewpoint of sentiment}.
For example, as illustrated in Figure~\ref{fig:fig1}, the two sentences ``The food is delicious.'' and ``The atmosphere of the restaurant is good.'' are very different in terms of semantics, but are similar with regards to sentiment.

For further improvement, recent studies investigated the approaches toward building \textit{sentiment-aware} PLMs: They introduced sentiment information in words, phrases, and sentences and designed sentiment-related pre-training objectives into PLMs. Notable examples include phrase-level polarity prediction adopted by SentiBERT \citep{yin2020sentibert}, combined word-level and sentence-level polarity prediction proposed by SentiLARE \citep{ke2019sentilare}, word-level and sentence-level polarity and emoticon prediction performed by SentiX \citep{zhou2020sentix}, and prediction of the replaced sentiment words and sentence-level contrastive learning in SentiWSP ~\citep{fan-etal-2022-sentiment}.

Even though they have achieved great performance in their downstream sentiment analysis tasks, quantitative evaluation of the quality of their constructed sentiment representations has been largely neglected. 
We argue that the quality of sentiment representations should be guaranteed because, without it, the low-quality representations of a backbone model may lead to the risk of poor \textit{few-shot} performance when applied to other sentiment-related domains with limited labeled data~\citep{tian2020rethinking,rizve2021exploring}. In light of the data scarcity in many real-world scenarios~\citep{gunel2021supervised}, it is crucial to consider few-shot learning rather than just focusing on fine-tuning performance, which would be mainly due to the supervision of sufficient fine-tuning labels.

This paper presents a novel sentiment-aware pre-training framework that ensures high-quality sentiment representations. First, we propose \textbf{S}entiment-\textbf{g}uided \textbf{T}extual \textbf{S}imilarity (\textbf{SgTS}), a novel metric that quantitatively evaluates the quality of sentiment representations by measuring the extent of equivalence in sentiment polarity between two provided sentences.
We then propose \textbf{SentiCSE}: Sentiment-aware Contrastive Sentence Embedding, which is highly focused on constructing sentiment representations by utilizing sentiment-related linguistic knowledge, towards achieving better performance on downstream sentiment-related tasks in label-scarce scenarios. SentiCSE has the following two pre-training objectives. First, the word-level objective aims to reconstruct sentences where each predicted word has the same sentiment polarity with the masked token, towards a better understanding of sentiment in text rather than semantics. Second, the sentence-level objective utilizes a quadruple of sentences (two positives and two negatives), and puts it to the supervised contrastive learning objective, where the sentences with the same polarity attract each other, and those with different polarity, which can be considered as negatives, repel each other. Through our extensive experiments under both the full label (linear probing) and the few-shot settings, we confirm that SentiCSE not only achieves state-of-the-art performance but also exhibits the best quality of sentiment representations using much less data.

\section{SgTS} 
\begin{figure}[t]
    \centerline{\includegraphics[width= 0.47\textwidth]{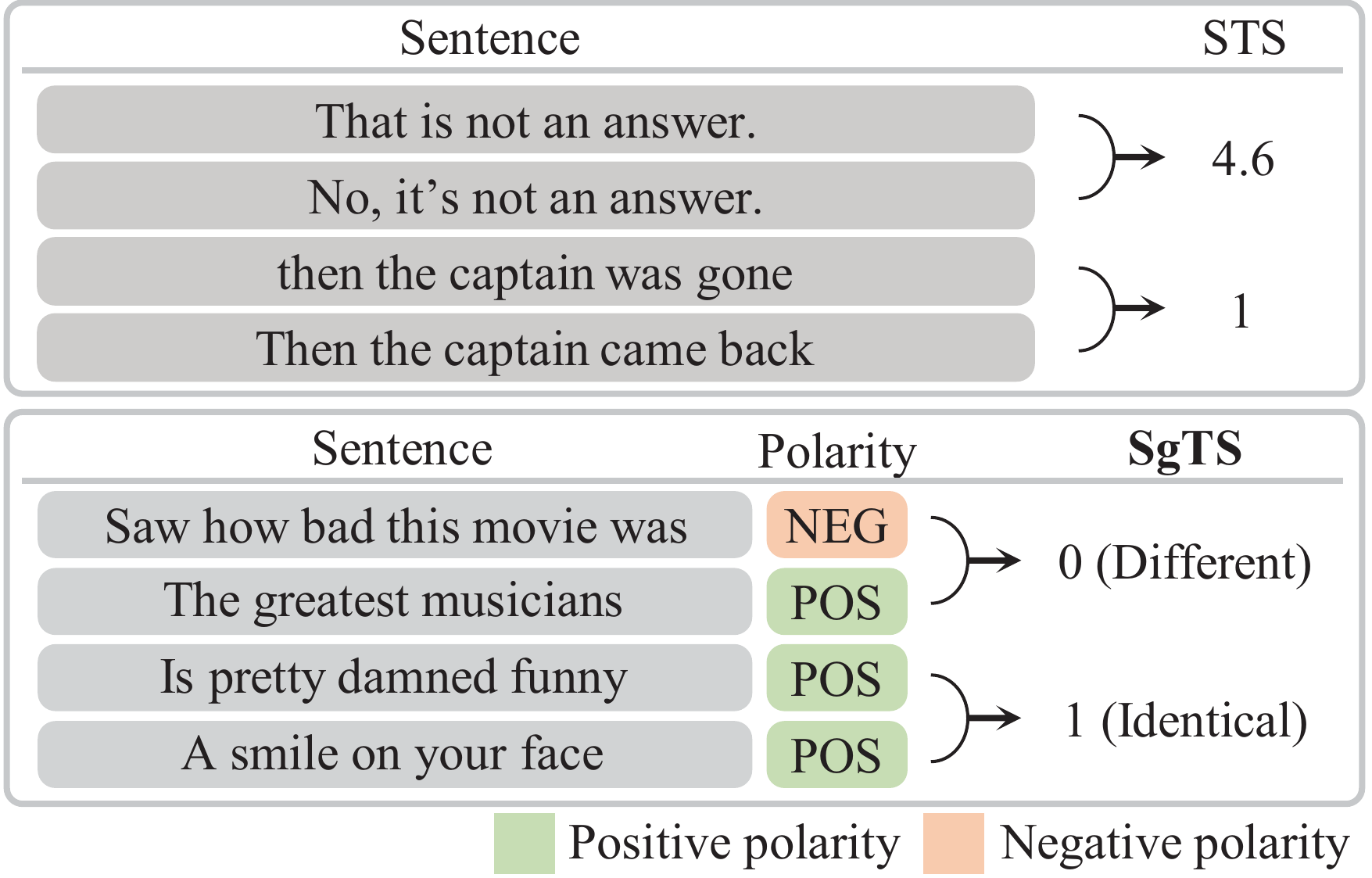}}
    \caption{Comparison of STS and our SgTS. STS measures similarity of two sentences based on contextual semantics while SgTS judges similarity based on their sentiment polarities.}
   \label{fig:squirrel}
   \vspace{-2mm}
\end{figure}

\noindent To the best of our knowledge, there has been no method to quantitatively evaluate the quality of sentiment representations. Traditionally, \textit{Semantic Textual Similarity} (STS)~\citep{agirre2012semeval,agirre-etal-2013-sem,agirre-etal-2014-semeval,agirre2015semeval,agirre2016semeval,cer-etal-2017-semeval,marelli-etal-2014-sick} has been a popular task that measures the degree of semantic equivalence between two sentences and has been employed by the original PLMs to measure the quality of their contextualized representations by using Spearman's correlation~\citep{reimers-gurevych-2019-sentence}. However, directly applying STS to evaluate the quality of sentiment representations is not feasible because the textual similarity between two sentences in terms of semantics is quite different from the similarity in terms of sentiment.

To fill this research gap, we propose an SgTS (\textbf{S}entiment-\textbf{g}uided \textbf{T}extual \textbf{S}imilarity) metric that can evaluate the quality of sentiment representations. We first design an SgTS task, inspired by the STS task, which aims to measure the sentiment similarity of two given sentences. Figure~\ref{fig:squirrel} illustrates our SgTS task. We assigned a ground-truth similarity score of 0 (different) or 1 (identical) to each pair of two sentences based on their sentiment polarities. In this way, we can automatically construct an SgTS benchmark (i.e., a corpus of sentence pairs with the similarity labels of 0 or 1) based on any dataset with the sentiment labels on sentences. Finally, the quality of sentiment representation is measured by the Spearman correlation between the ground-truth similarity scores (0 or 1) and the predicted scores computed by cosine similarity between two given sentence embeddings.  

\begin{figure*}[ht!]
    \centerline{\includegraphics[width=\textwidth]{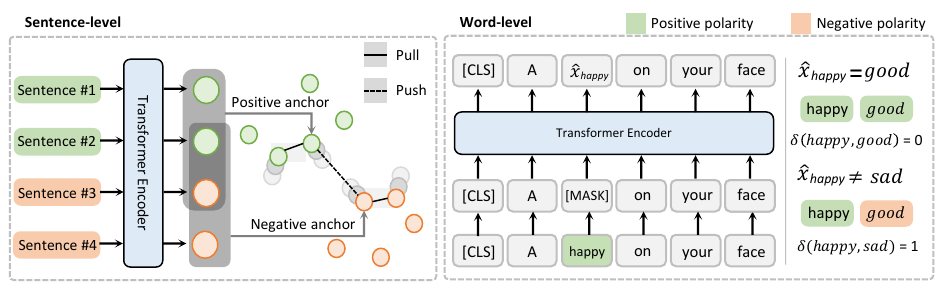}}
    \caption{The overview of SentiCSE. In the sentence-level objective, we take two anchors, one for `positive' polarity and the other for `negative' polarity. We then encourage the sentiment representations to be more close to the corresponding sentences belonging to the same polarity, and to be far from the corresponding sentences associated with different polarities. In the word-level objective, our model tries to predict the masked words as in conventional MLM. But the error signal is dropped if the model matches the polarity of the predicted word with the masked word, even though the word prediction was not correct.}
   \label{fig:sentiment models embedding}
\end{figure*}

\section{SentiCSE}
This section presents SentiCSE, an effective framework for learning high-quality sentiment representations. It borrows the encoder of backbones such as BERT or RoBERTa. Its overall framework is illustrated in Figure~\ref{fig:sentiment models embedding}. SentiCSE consists of two sentiment-aware pre-training objectives: the Word-level objective that recovers the sentiment polarity of masked tokens and the Sentence-level objective based on supervised contrastive learning. The main idea of our design is to allow SentiCSE to improve the quality of its sentiment representation by leveraging the characteristics of sentiment polarity.

\subsection{Sentiment-aware Word-level Objective}
Formally, let $X=\{x_1, x_2, \cdots, x_{n}\}$ be a text sequence of length $n$, where each $x_i$ is a word in the Byte-Pair Encoding vocabulary~\citep{radford2019language} with the size of 50,265. In the original masked language model (MLM), a certain percentage of words in a given sentence is randomly masked and then predicted by the model~\citep{devlin-etal-2019-bert,clark2020electra}. In our work, we selectively mask the sentiment words included in the SentiWordNet lexicon~\citep{baccianella-etal-2010-sentiwordnet}. More specifically, within the word listed in SentiWordNet, we refer to its sentiment scores (i.e., \textit{positivity} and \textit{negativity}). According to the two scores, we acquire the word's polarity  based on the following polarity function:
\begin{equation}
\resizebox{0.5\textwidth}{!}{%
    $\mathcal{S}(x)$ = $\begin{cases}
    +,    \quad\text{if $S_{pos}(x) > 0$ and $S_{neg}(x)==0$}, \\
    -,    \quad\text{if $S_{pos}(x)==0$ and $S_{neg}(x)>0$}, \\
    \text{\textit{multi}}, \text{if $S_{pos}(x)>0$ and $S_{neg}(x)>0$},
    \end{cases}$
}
\end{equation} 
where $S_{pos}(x)$ and $S_{neg}(x)$ indicate the positivity and negativity scores of the given word $x$, respectively. Among the words in a sentence, only the words having either `$+$' (positive) or `$-$' (negative) polarity are candidate tokens for masking. We do not mask the words with `\textit{multi}' (multi-polarity) because such words are positive in some sentences and negative in some other sentences, which would make training unstable. This setting helps the training to focus more on sentiment-related information. We mask 10\% of the candidate tokens; we empirically found that the masking ratio of 10\% balances the trade-off between providing sufficient sentiment information to learn good representations and promoting training efficiency.

After generating a corrupted sentence, our model is trained to predict the masked words with the same sentiment as the original words while also predicting the masked words as original words. Formally, let $\Tilde{X} = \{\Tilde{x}_1, \Tilde{x}_2, \cdots, \Tilde{x}_{n}\}$ be the corrupted version of $X$. The encoder encodes $\Tilde{X}$ to contextualized representations $ H =\{h_1, h_2, \cdots, h_{n}\}$. Let $k$ denote the index of the masked words; then each $h_k$ is fed to a SoftMax layer: $P_\theta  (m_{k}|M) = Softmax(W\cdot h_k+b)$ where $\theta$ denotes the model parameters.

After then, we replace the masked word $\Tilde{x}_{k}$ with the predicted word $\hat{x}_{k}$ that has the highest softmax probability. We then introduce an indicator function $\delta(x_k,\hat{x}_{k})$ that outputs 0 if the polarities of the two given words $x_k$ and $\hat{x}_{k}$ are identical, and 1 if they are different, based on SentiWordNet:
\begin{equation}
    \delta(x_k,\hat{x}_{k}) = \begin{cases}
    0,&\text{if $\mathcal{S}(x_k)==\mathcal{S}(\hat{x}_{k})$} \\
    1,&\text{otherwise.}
    \end{cases}
\end{equation}
This indicator function is applied to the cross-entropy loss for a single sentence $X$ is:
\begin{equation}
\mathcal{L}_{w}^{X} = \sum_{k\in \mathcal{I}}-logP_\theta(x_k|\Tilde{X})\cdot \delta(x_k, \hat{x}_{k}),
\end{equation}
where $\mathcal{I}$ denotes the set of indices of the masked words. Then the training objective for a batch is:
\begin{equation}
\mathcal{L}_{w}= \sum_{i=1}^{N}\mathcal{L}_{w}^{X_i},
\end{equation}
where $N$ is the batch size.

By doing so, the loss caused by word prediction that matches the polarity of the original word is removed, even if the words themselves are not matched. This objective makes the model focus more on understanding linguistic knowledge in terms of sentiment.

\subsection{Sentiment-aware Sentence-level Objective}
Here, we adopt the contrastive learning framework demonstrated in SimCSE~\citep{gao-etal-2021-simcse}. Especially, we choose \textit{supervised} contrastive learning because it can explicitly define which examples should be attracted or should be repelled, leveraging negatives. We believe that the sentiment analysis task benefits from negatives, which can be easily chosen based on the sentiment polarities of sentences.

Formally, we compose a contrastive example with a quadruple of sentences $q_i$ : ($p_{i},p_{i}^+,n_{i},n_{i}^+$), where $p_{i}$ and $p_{i}^+$ are two different positive sentences and $n_{i}$ and $n_{i}^+$ are two different  negative sentences.
We then define two objectives, $\mathcal{L}_{pos}^{q_i}$ and $\mathcal{L}_{neg}^{q_i}$, each pulling the given two positive sentences and pushing the two negative sentences close to each other, respectively:
\begin{equation}\label{Losspos}
\resizebox{0.43\textwidth}{!}{%
$-\text{\footnotesize{log}}\frac{e^{sim(p_{i},p_{i}^+)/\tau }}{e^{sim(p_{i},p_{i}^+)/\tau}+\sum_{j=1}^{N}(\alpha \cdot e^{sim(p_{i},n_{j})/\tau})}$,
}
\end{equation}
\begin{equation}\label{Lossneg}
\resizebox{0.43\textwidth}{!}{%
$-\text{\footnotesize{log}}\frac{e^{sim(n_{i},n_{i}^+)/\tau }}{e^{sim(n_{i},n_{i}^+)/\tau}+\sum_{j=1}^{N}(\alpha  \cdot e^{sim(n_{i},p_{j}^+)/\tau})}$,
 }
 \end{equation}
where $sim(\cdot)$ computes the cosine similarity between the embeddings of two given sentences, $\alpha$ indicates the weight on negative samples, and $\tau$ stands for temperature. The hyperparameters $\alpha$ and $\tau$ control the effectiveness of negative samples. Our sentence-level objective for a batch $\mathcal{L}_{s}$ simply combines the two above losses by:
\begin{equation}\label{loss term}
\mathcal{L}_{s} = \frac{1}{N}\sum_{i=1}^{N}(\mathcal{L}_{pos}^{q_i} + \mathcal{L}_{neg}^{q_i}).
\end{equation}
With this objective, SentiCSE strives to learn high-quality representations by pulling neighbors sharing similar sentiment together and pushing apart non-neighbors with different sentiment.

The final objective function of our SentiCSE can be denoted as:
\begin{equation}\label{loss term}
\mathcal{L}_{SentiCSE} = \lambda_w \cdot \mathcal{L}_{w} + \mathcal{L}_{s},
\end{equation}
where $\lambda_w$ controls the importance of the word-level loss. We empirically choose the values of $\lambda_w$ as 0.15. Guided by the combined word-level and sentence-level objectives, SentiCSE learns representations that are more reflective of the sentiment polarity of the sentences.

\section{Experimental Settings}\label{exp:Training Settings}
We conducted extensive experiments to evaluate our SentiCSE. This section summarizes the implementation details of our experiments.

\subsection{Dataset} \label{method:choice dataset}
We employed the following four datasets that include sufficient sentiment words: \textbf{IMDB}~\citep{maas-etal-2011-learning}, \textbf{SST-2} ~\citep{socher-etal-2013-recursive}, \textbf{Yelp-2}~\citep{NIPS2015_250cf8b5}, \textbf{Amazon}~\citep{ni-etal-2019-justifying}, and \textbf{MR}~\citep{pang-lee-2005-seeing}. All the datasets include review sentences collected from various domains (movie, restaurant, dental clinic, etc.), whose sentiment labels are given by human experts or automatically judged based on the corresponding review scores. Table~\ref{table:dataset static} shows their statistics.

\begin{table}[t] \
\centering
\resizebox{0.47\textwidth}{!}
{%
\begin{tabular}{l|rrrrr}
\hline
\textbf{Dataset}  & \textbf{IMDB} & \textbf{SST-2} & \textbf{Yelp-2}  & \textbf{Amazon}  &  \textbf{MR}\\ \hline \hline
Avg. length          & 302	        & 14	    & 176       &42& 27        \\
\% of SentiWords     & 6.3\%     	& 8.3\%     & 5.1\%     &6.0\%& 9.9\%     \\  
\# train             & 25,000	    & 67,349	& 560,000   &160,000& 8,530     \\
\# valid             &	 -         	& 872	    & -         &4,000& 1,066     \\
\# test              & 25,000	    & 1,821	    & 38,000    &4,000& 1,066     \\
\hline
\end{tabular}%
}
\caption{Data statistics.}
\label{table:dataset static}
\end{table}

\definecolor{Gray}{gray}{0.91}
\begin{table}
\centering
\resizebox{0.47\textwidth}{!}{%
\begin{tabular}{c|c|ccccc|c}
\Xhline{3\arrayrulewidth}
\multicolumn{1}{c|}{\multirow{2}{*}{\textbf{Model}}} & \multicolumn{1}{c|}{\multirow{2}{*}{\textbf{Backbone}}} & \multicolumn{5}{c|}{\textbf{Pre-trained dataset}} & \multicolumn{1}{c}{\multirow{2}{*}{\textbf{\# Sentences}}}\\ \cline{3-7}
\textbf{} & & \textbf{Wiki} & \textbf{Amazon} & \textbf{Yelp} & \textbf{SST} & \textbf{MR} & 
 \\ \hline
SentiBERT   & BERT      &  &  &   & $$ \checkmark $$ &  & 0.067M \\
SentiX      & BERT      &  & $$ \checkmark $$  & $$ \checkmark $$  &  &  & 240M \\
SentiLARE   & RoBERTa   &  &  &  $$ \checkmark $$  &  &  & 6.7M \\
SentiWSP    & ELECTRA   &  $$ \checkmark $$ &  &  &  &  & 0.5M \\
SentiCSE    & RoBERTa   &  &  &  &  & $$ \checkmark $$ & 0.008M \\
\Xhline{3\arrayrulewidth}
\end{tabular}}
\caption{The default pre-training corpora with their number of sentences and backbones adopted by each method.}
\label{tab:pre-trained datasets}
\end{table}

\subsection{Baselines}
We employed the four sentiment-aware PLMs including \textbf{SentiBERT}~\citep{yin2020sentibert}, \textbf{SentiX}~\citep{zhou2020sentix}, \textbf{SentiLARE}~\citep{ke2019sentilare}, and \textbf{SentiWSP}\footnote{Actually, SentiWSP adopts quite different configurations from the other baselines, such as the use of ELECTRA, Wiki dataset, and the unsupervised contrastive learning. And its performance is not on par with the other baselines (SgTS: Avg. 0.02, linear probing: Avg. 85.76, 1-shot: Avg. 56.23, 5-shot: Avg. 63.67). Hence, it is not compared in most of our comparative results.}~\citep{fan-etal-2022-sentiment}. 
We also employed supervised \textbf{SimCSE}~\citep{gao-etal-2021-simcse}, \textbf{BERT}~\citep{devlin-etal-2019-bert}, and \textbf{RoBERTa}~\citep{liu2019roberta} as baselines. The pre-training corpora and the backbones adopted in each baseline can be found in Table~\ref{tab:pre-trained datasets}. BERT and RoBERTa were pre-trained on the Wiki corpus~\citep{devlin-etal-2019-bert} same with SentiWSP; supervised SimCSE was trained on the NLI dataset~\citep{bowman-etal-2015-large, williams-etal-2018-broad} transformed for contrastive learning.

For a fair comparison, there can be two perspectives on what should be selected as the pretraining data and backbone model for each compared method. The first idea is to follow each baseline's own configuration; more specifically, we use the pretraining dataset and backbone model chosen by the authors of the respective baseline. This protocol has been adopted by most of the baseline papers (SentX, SentiLARE, and SentiWSP), in the belief that it would be inappropriate to use a different backbone model or pretraining corpus which were not selected by the authors of each baseline. In contrast, the second perspective is that a fair evaluation requires the use of the same pretraining data and backbone models. We try our best to conduct the comparative experiments by following both two perspectives, though the second perspective is harder to implement because the baselines generally adopt different pretraining datasets and backbones with each other.

\subsection{Pre-training Environment}
We briefly explain how we pre-trained SentiCSE. We chose MR as the default pretraining dataset, but other dataset can be used. For each positive sentence contained in a training set, we composed a quadruple of sentences by randomly sampling another positive sentence and two negative sentences from the training set (in the case where MR is adopted as a pretraining dataset, 4,265 quadruples will be generated). For the hyperparameters, we fixed the maximum sentence length as 128, the number of layers (Transformer blocks) as 12, and the embedding dimension as 768. The learning rate and the batch size were set to 1e-5 and 64, respectively. Under this setting, if we use MR, the pre-training takes only 3 hours, which proceeds 20,000 steps with four NVIDIA A30 GPUs. For the validation step, we constructed the SgTS benchmark (if MR is used, the SgTS-MR benchmark is composed of 429 sentence pairs with 0 or 1 labels, based on the 872 sentences included in MR's validation set). For every 500 step, we evaluated the SgTS performance to get the best checkpoint (In the case where MR is used, we obtained the best checkpoint at 14,000 step).

\begin{figure*}[t]
    \centerline{\includegraphics[width=\textwidth]{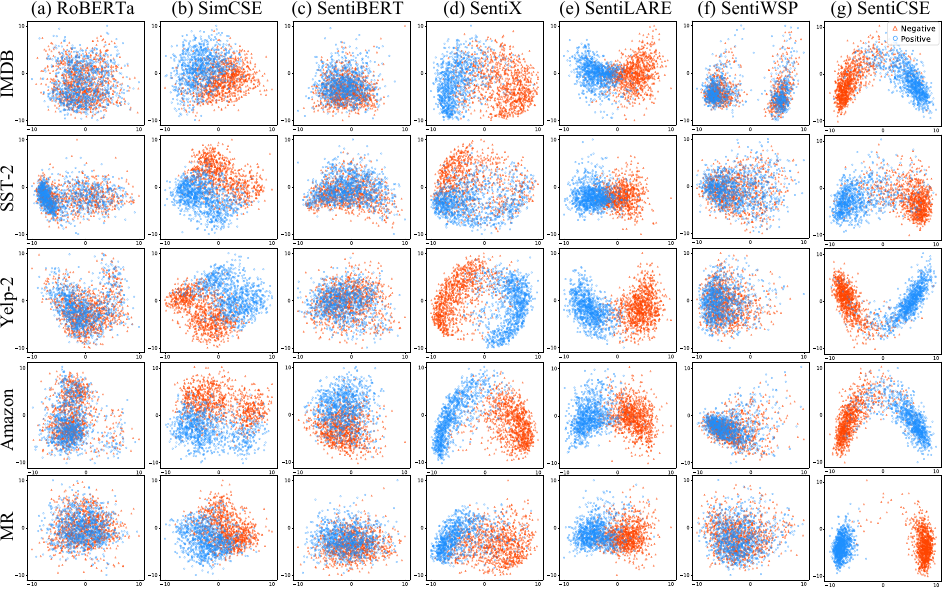}}
    \vspace{-2mm}
    \caption{Visualized embeddings of the sentences included in each {dataset}. Each model is pre-trained on its own pretraining data, specificed in Table~\ref{tab:pre-trained datasets}. PCA is used for visualization.}
   \label{fig: sentiment models embedding}
\end{figure*}

\subsection{Details in Downstream Tasks}
To evaluate the downstream performance of SentiCSE, we employed the {linear probing} and {few-shot} settings. Linear probing only trains the linear classification layer on top of the fixed input representations. We followed the default configurations for linear probing, which can be found in the SentEval toolkit~\citep{conneau-kiela-2018-senteval}. Since the configurations for IMDB and Yelp-2 are not available, we borrowed the settings for SST-2. Since IMDB and Yelp-2 do not provide validation data, we randomly sampled 10\% of their training sets to construct the validation sets.

In contrast to linear probing supervised by the entire training sentences, the few-shot setting assumes a data-scarce situation. This setting limits the number of accessible samples for each class (positive or negative) to $K$. For example, in the 5-shot setting, we randomly sampled 10 sentences (5 for each class) without replacement. Among the rest training data, we randomly sampled 500 examples for the validation, which was conducted every 20 steps with the early stop patience value as 5. Since the number of training examples is extremely small, the random seed may have a lot of influence on performance. To mitigate this issue, we repeated 10 experiments with 10 random seeds and then reported the average of the results~\citep{gunel2021supervised}. We tested SentiCSE on the 1-shot and 5-shot settings.

\begin{table}[t]
\centering
\resizebox{0.47\textwidth}{!}{%
\begin{tabular}{l|ccccc|c}
\Xhline{3\arrayrulewidth}
\multirow{2}{*}{\textbf{Model}} & \multicolumn{5}{c}{\textbf{SgTS}}   \\   \cline{2-7}
 & \textbf{IMDB} & \textbf{SST-2} & \textbf{Yelp-2} & \textbf{Amazon} & \textbf{MR} & \textbf{Avg.}\\ \hline
BERT            & 0.01 & 0.08 & 0.09 & 0.15 & 0.07 & 0.06 \\
$\text{SimCSE}_{BERT}$  & 0.16 & 0.13 & 0.24 & 0.19& 0.13 & 0.18 \\
SentiBERT   & 0.13 & 0.17$^*$ & 0.12 & 0.09 & 0.18 & 0.14 \\
SentiX      & 0.62 & 0.48 & 0.77$^*$ & 0.52$^*$ & \textbf{0.39} & 0.56 \\
$\text{SentiCSE}_{BERT}$& \textbf{0.64} & \textbf{0.72} & \textbf{0.76} & \textbf{0.37} & 0.63$^*$ & \textbf{0.62} \\
\hline
\hline
Roberta & 0.06 & 0.05 & 0.06 & 0.02 & 0.04 & 0.06 \\
$\text{SimCSE}_{RoBERTa}$  & 0.21 & 0.11 & 0.26 & 0.20 & 0.19 & 0.19 \\
SentiLARE   & 0.48 & 0.38 & 0.65$^*$ & 0.36 & \textbf{0.57} & 0.46 \\
\rowcolor{Gray}
$\text{SentiCSE}_{RoBERTa}$       & \textbf{0.77} & \textbf{0.72} & \textbf{0.82} & \textbf{0.56} & 0.69$^*$ & \textbf{0.71} \\
\Xhline{3\arrayrulewidth} 
\end{tabular} %
}
\caption{Representation quality measured by our SgTS task. We employ Spearman's correlation with the ``all'' setting. We divided the compared methods into two groups: the BERT-based models and RoBERTa-based ones. The highest scores for each data are highlighted in bold face. For each data column, the scores with * indicate the results when the same dataset is used for pretraining and a downstream task.} 
\label{tab:SgTS}
\end{table}

\section{Results and Analyses} \label{exp:Main Results}
This section reports the results of our extensive experiments.

\begin{figure}%
\subfloat[SgTS-Accuracy]{{\includegraphics[width=0.24\textwidth ]{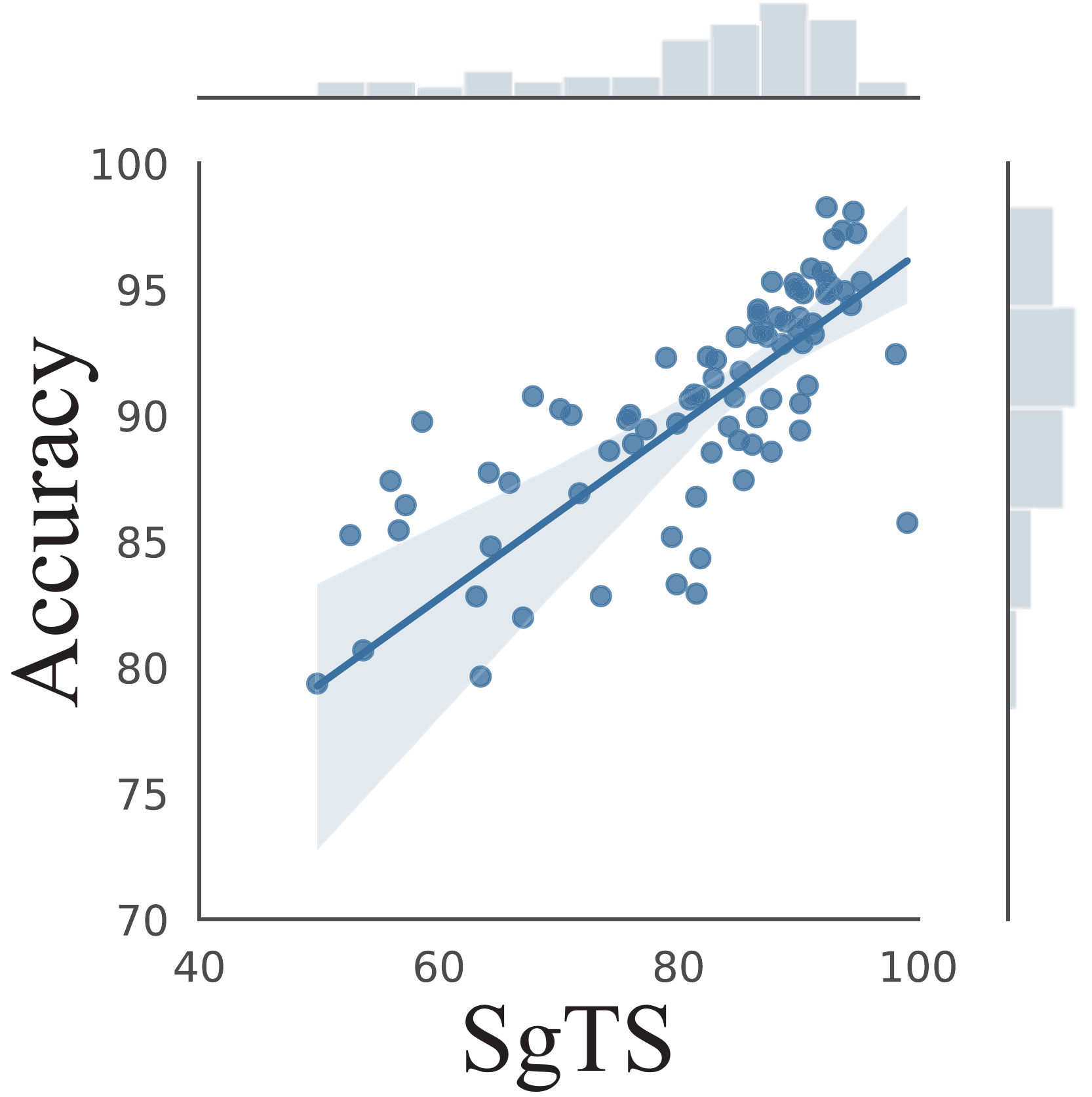} }}%
\subfloat[STS-Accuracy]{{\includegraphics[width=0.24\textwidth ]{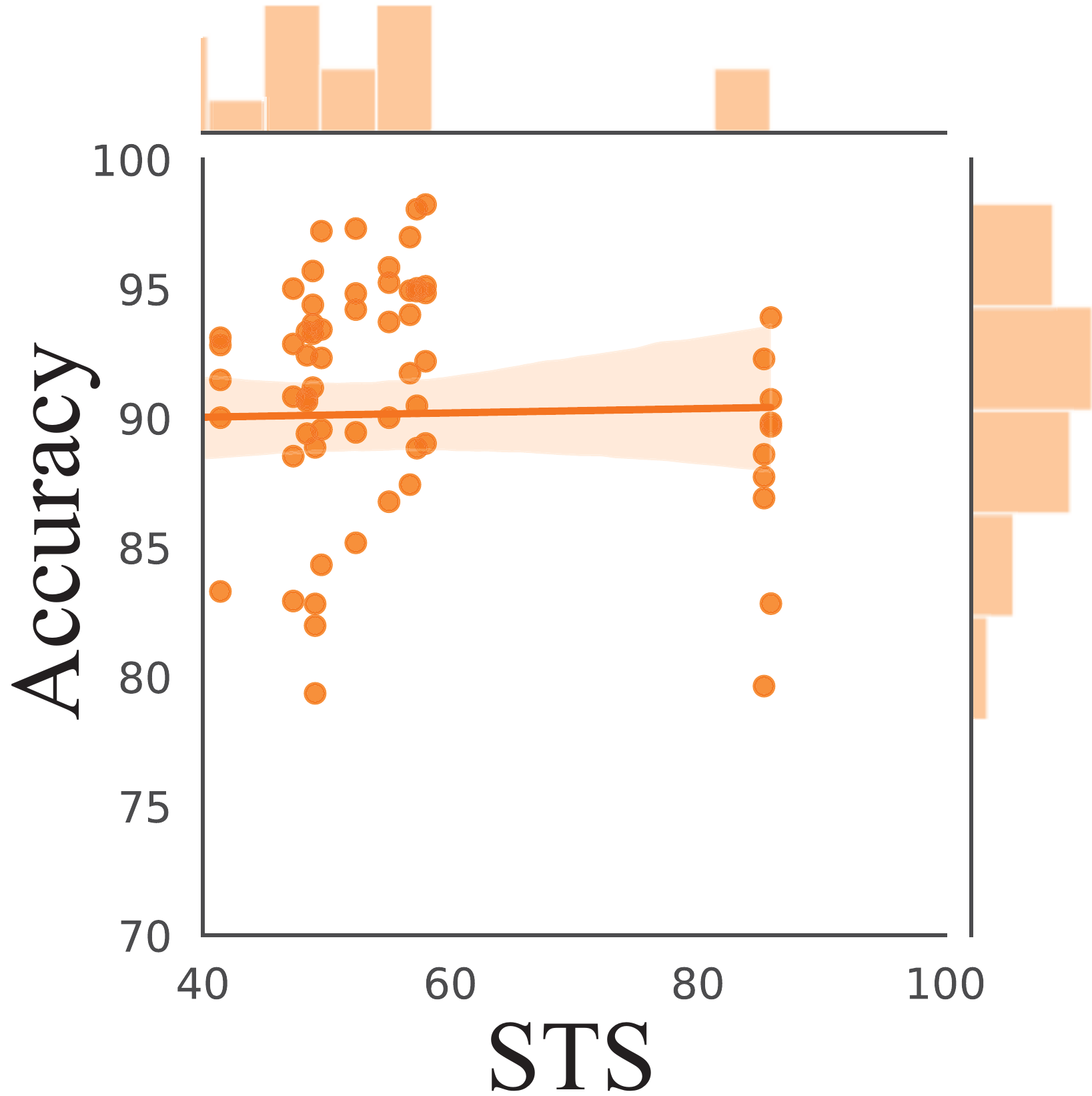} }}%
\hfill
\caption{(a): Correlation between our SgTS and few-shot accuracy. (b): Correlation between STS and few-shot accuracy. We can see there is a strong correlation between our SgTS and the accuracy of sentiment analysis ($\rho=0.7$ with $p$-value $< 0.01$) while there does not seem to be a correlation between STS and accuracy.}
\vspace{-5mm}
\label{fig:SgTS_STS_corr}%
\end{figure}

\subsection{Quality of Representations (SgTS)} \label{exp_sub:Representation Analysis}
First, we qualitatively and quantitatively evaluates the quality of sentiment representations of each model. Figure~\ref{fig: sentiment models embedding} compares each model's visualization of sentiment representation on each downstream dataset: for each data, we let each model encode sentences in the dataset to 768-dimensional embedding vectors. We then projected them to 2D space via PCA (principal component analysis). We observed that SentiX, SentiLARE, and SentiCSE exhibit better quality of representations, which indicates their representations contain much more sentiment information than the other models (SentiWSP, RoBERTa and SimCSE) pre-trained on a dataset that is not related to sentiment analysis. We observed that our SentiCSE exhibits better representations than the baselines because it makes the distance between the two clusters (positive sentence embeddings and negative sentence embeddings) far enough.

In addition, Table~\ref{tab:SgTS} reports the representation quality in terms of the proposed SgTS metric, measured by Spearman correlation between the Cosine similarity between two sentence embeddings and the corresponding ground-truth labels (0: `different' or 1: `identical'). We observed that our SentiCSE exhibits the highest value, which indicates that two sentences with the same sentiment polarity are close to each other in the embedding space, and far from each other if their polarities are different. We also observed that SentiX and SentiLARE exhibit relatively high scores on Yelp-2. This also shows the effectiveness of our SgTS as an evaluation method for the quality of sentiment representation, because Yelp-2 is the pretraining dataset of SentiX and SentiLARE. In addition, the sentiment-aware PLMs except our SentiCSE show relatively low score on MR and SST-2, where the average length of the sentences is shorter than that of the other dataset. This might imply that the previous sentiment-aware PLMs rely on sufficient semantic information for constructing sentiment representations. 

\begin{table*}[t] 
\centering
\resizebox{1\textwidth}{!}{%
\begin{tabular}{lcccccccccccc}
\Xhline{2\arrayrulewidth}
 \multirow{2}{*}{\textbf{Model}} & 
 \multicolumn{5}{c}{\textit{\textbf{1-shot accuracy}}} &&\multicolumn{5}{c}{\textit{\textbf{5-shot accuracy}}} \\
 \cline{2-12}& \multicolumn{1}{c}{\textbf\textit{IMDB}} & SST2 & Yelp-2 &Amazon& MR  & & IMDB & SST2 & Yelp-2 &Amazon& MR  \\ \hline
BERT                    & 52.08 & 50.26 & 56.76 &52.98 & 52.24          && 54.02 & 54.26 & 62.64 & 58.10 & 54.38 \\
$\text{SimCSE}_{BERT}$  & 54.08 & 61.74 & 66.20 & 60.92 & 61.64         && 71.26 & 66.82 & 81.58 & 73.58 & 67.16 \\
SentiBERT               & 51.40 & 55.60$^*$ & 59.64 & 54.90 & 54.88      && 57.76 & 64.84$^*$ & 70.20 &67.02& 64.90 \\
SentiX & 74.64 & 64.96 & 87.66$^*$ & 86.14$^*$& \textbf{65.06} && \textbf{83.68} & 72.32 & 93.40$^*$ & 92.32$^*$ & \textbf{76.68} \\
\rowcolor{Gray}
$\text{SentiCSE}_{BERT}$ & \textbf{76.08} & \textbf{87.88} & \textbf{81.62} &\textbf{82.24}& 85.82$^*$ && 81.84& \textbf{93.26} & \textbf{87.64} & \textbf{84.82} & 86.14$^*$ \\
\hline
\hline
RoBERTa                     & 52.00 & 54.54 & 56.56 & 52.84 & 53.82       && 60.30 & 49.80 & 72.42 & 64.58 & 56.78 \\
$\text{SimCSE}_{RoBERTa}$   & 59.04 & 61.06 & 68.44 & 58.40 & 61.72       && 74.72 & 68.08 & 86.62 & 75.14 & 71.56 \\
SentiLARE                   & 70.20 & 74.26 & 87.00$^*$ &84.58& \textbf{68.68} && 87.18 & 80.10 & 93.28$^*$ &\textbf{91.06}& \textbf{82.34} \\
\rowcolor{Gray}
$\text{SentiCSE}_{RoBERTa}$ & \textbf{82.64} & \textbf{92.92} & \textbf{89.72} & \textbf{89.04} & 87.38$^*$ && \textbf{88.12} & \textbf{94.50} & \textbf{92.08} &90.40& 88.00$^*$ \\
\Xhline{2\arrayrulewidth}
\end{tabular} %
}
\caption{Comparative results on the 1-shot and 5-shot settings on each downstream dataset. In this table, we divided the compared methods into two groups: the BERT-based models and RoBERTa-based ones. Then we separately compared them within each group. For each data column, the accuracy with * indicate the results when the same dataset is used for pretraining and a downstream task. We do not highlight such cases in bold face. Instead, we highlight the best accuracy achieved by the models pre-trained on a different dataset. SentiCSE achieves an impressive few-shot performance in most cases, owing to its high-quality sentiment representations.}
\label{table:few-shot}
\end{table*} 
\definecolor{Gray}{gray}{0.91}
\begin{table}[h!]
\centering
\resizebox{0.47\textwidth}{!}{%
\begin{tabular}{lccccc} 
\Xhline{3\arrayrulewidth}
\textbf{Model} & \multicolumn{1}{l}{\textbf{IMDB}} & \multicolumn{1}{l}{\textbf{SST-2}} & \multicolumn{1}{l}{\textbf{Yelp-2}} & \multicolumn{1}{l}{\textbf{Amazon}}& \multicolumn{1}{c}{\textbf{MR}} \\ 
\hline
BERT                &   85.25   &   85.44   &   89.75   & 86.44 &   80.68  \\
$\text{SimCSE}_{BERT}$ &   86.91   &   87.73   &   92.29   & 88.60 &  79.64  \\
SentiBERT       &   87.40   &   $90.25^*$   &   90.76   & 87.33 & 84.80  \\
SentiX              &   \textbf{94.20}   &   89.45   &   $97.33^*$   & $94.82^*$ & \textbf{85.18}\\
\rowcolor{Gray}
$\text{SentiCSE}_{BERT}$ &   90.63  &   \textbf{95.30}   &   \textbf{93.12}   & \textbf{89.93} & $85.74^*$  \\ \hline \hline
RoBERTa         &   82.82   &   79.36   &   88.87   & 81.98 &   50.38 \\ 
$\text{SimCSE}_{RoBERTa}$&  90.73   &   89.68   &   93.89   &  89.82 & 82.83 \\
SentiLARE       &   \textbf{94.84}   &   92.20   &   $98.26^*$   & \textbf{95.10} & \textbf{89.02} \\ 
\rowcolor{Gray}
$\text{SentiCSE}_{RoBERTa}$ &   94.03  &   \textbf{95.18}   &   \textbf{95.86}   & 93.69 & $89.49^*$      \\ 
\Xhline{3\arrayrulewidth}
\end{tabular}}
\caption{Comparison of linear probing performance (i.e., a full resource setting).}
\label{tab: linear probing}
\end{table}

Finally, Figure~\ref{fig:SgTS_STS_corr} clearly depicts the strong correlation between the few-shot accuracy (this will be reported in the next subsection) and our SgTS metric. We computed the Pearson Correlation Coefficient among the SgTS scores and the corresponding few-shot accuracy scores, and got $\rho=0.7$ with $p$-value $< 0.01 $, which indicates that high-quality representation guaranteed by SgTS results in the high few-shot performance. Such strong relationship implies that our SgTS serves as a meaningful quantitative metric for measuring the quality of sentiment representation. In contrast, the relationship between the traditional STS and the accuracy of sentiment analysis does not appear to be significant. This result sheds a new light on how to pre-train a language model for better sentiment analysis.

\subsection{Performance of Downstream Tasks}  \label{exp_sub:Few shot Learning}


First, Table~\ref{table:few-shot} reports the comparative results on the few-shot setting. We observed that SentiCSE achieved significant performance improvement. A notable point is that SentiCSE even outperforms SentiX and SentiLARE on the Amazon and Yelp-2 dataset, which are their pretraining datasets; it also beats SentiBERT by a wide margin on SST2, on which SentiBERT was pre-trained.

Next, Table~\ref{tab: linear probing} shows results on the linear probing setting that provides full resource. Again, SentiCSE exhibits impressive perfornamce: SentiCSE outperforms SentiBERT on SST2 where SentiBERT was pre-trained. SentiCSE slightly underperforms SentiLARE on IMDB. SentiCSE also underperforms SentiLARE and SentiX on Yelp-2 and Amazon, however, considering that Yelp-2 and Amazon are their pretraining datasets, we believe that the results still support the effectiveness of our work.

Another noteworthy point is that the baselines were pre-trained on very large corpus such as Yelp, which involved significant computational resources and time. In contrast, SentiCSE was pre-trained on MR, which contains only 1.52\% of the training data of Yelp-2. As a result, SentiCSE requires much less resources in terms of data, computation, and cost than the baselines for both pre-training and downstream tasks, but is able to learn high-quality sentiment representations and provides comparable downstream performance between them.

Finally, we try to compare SentiCSE with the baselines using the same pretraining dataset and backbone model. For this experiment, we implemented various versions of SentiCSE using different backbones (BERT and RoBERTa) and different pretraining dataset (SST2, Yelp2, and Amazon). Table~\ref{table:additional_fewshot} reports the results. Again, we confirm that SentiCSE generally outperforms the baselines, even though SentiCSE gives up its best configuration (MR and RoBERTa) and is adapted to each baseline's best configuration.

\subsection{Further Analysis}  \label{exp:Ablation Study}
This subsection provides additional results from our ablation studies, hyperparameter analysis, and qualitative analysis.

\begin{table*}[t] 
\centering
\resizebox{1\textwidth}{!}{%
\begin{tabular}{clcccccccccccc}
\Xhline{2\arrayrulewidth}
 \multirow{2}{*}{\textbf{Dataset}}&\multirow{2}{*}{\textbf{Model}} & 
 \multicolumn{5}{c}{\textit{\textbf{1-shot accuracy}}} &&\multicolumn{5}{c}{\textit{\textbf{5-shot accuracy}}} \\
 \cline{3-13}& &\multicolumn{1}{c}{\textbf\textit{IMDB}} & SST2 & Yelp-2 &Amazon& MR  & & IMDB & SST2 & Yelp-2 &Amazon& MR  \\  \hline
 
\multirow{2}{*}{SST2}
&SentiBERT$^{\diamondsuit}$  & 51.40 & 55.60 & 59.64 & 54.90 & 54.88      && 57.76 & 64.84 & 70.20 &67.02& 64.90 \\
&SentiCSE$^{\diamondsuit}$   & \textbf{74.68 }& \textbf{91.82} & \textbf{82.00} &\textbf{81.24}& \textbf{86.94} &&\textbf{81.86}& \textbf{92.80} & \textbf{88.02} & \textbf{86.38} & \textbf{90.24} \\ \hline \hline

\multirow{4}{*}{Yelp2}
&SentiX$^{\diamondsuit}$    & 74.64 & 64.96 & 87.66 & 86.14$^*$& 65.06 && 83.68 & 72.32 & 93.40 & 92.32$^*$ & 76.68 \\
&SentiCSE$^{\diamondsuit}$  & 69.22 & 68.10 & 91.14 &85.48 & 63.76 && 84.24& 86.48 & \textbf{95.24} & 89.74 & 80.86 \\
&SentiLARE$^{\spadesuit}$   & 70.20 & 74.26 & 87.00 &84.58 & 68.68 && 87.18 & 80.10 & 93.28 &91.06 & 82.34 \\
&SentiCSE$^{\spadesuit}$    & \textbf{76.64} & \textbf{84.78} & \textbf{94.26} &\textbf{89.28}& \textbf{73.20} && \textbf{87.98} & \textbf{87.66} & 95.12 &\textbf{92.68} & \textbf{86.36}\\
\hline \hline

\multirow{2}{*}{Amazon}
&SentiX$^{\diamondsuit}$    & 74.64 & 64.96 & 87.66$^*$ & 86.14 & 65.06 && 83.68 & 72.32 & \textbf{93.40}$^*$ & 92.32 & 76.68 \\
&SentiCSE$^{\diamondsuit}$  & \textbf{75.16} & \textbf{78.56} & \textbf{91.16} &\textbf{93.16}& \textbf{78.02} &&\textbf{86.64}& \textbf{85.18} & 92.98 & \textbf{93.86} & \textbf{85.44} \\
\Xhline{2\arrayrulewidth}
\end{tabular} %
}
\caption{Comparative results on the 1-shot and 5-shot settings on each downstream dataset. In this table, we divided the compared methods into three groups in terms of pretraining data. Then we separately compared them within each group. Again, SentiCSE achieves meaningful few-shot performance in most cases. $\diamondsuit$: BERT-base models. $\spadesuit$: RoBERTa-based models.}
\label{table:additional_fewshot}
\end{table*} 
\begin{table}[t]
\centering
\resizebox{0.48\textwidth}{!}{%
\begin{tabular}{l|ccccc|c} 
\hline\
\textbf{Loss} & \multicolumn{1}{c}{\textbf{IMDB}} & \multicolumn{1}{c}{\textbf{SST-2}} & \multicolumn{1}{c}{\textbf{Yelp-2}} & \multicolumn{1}{c}{\textbf{Amazon}}& \multicolumn{1}{c|}{\textbf{MR}} 
 & \multicolumn{1}{c}{\textbf{Avg.}}  \\ \hline
 \quad $\mathcal{L}_{pos}$                  & 93.40 & 94.95 & 94.60 &92.13 & 86.59 & 92.33 
\\
 \quad $\mathcal{L}_{neg}$                  & 93.18 & 94.84 & 95.02 &92.38 & 87.99 & 92.68 
\\
 \quad $\mathcal{L}_{pos}+\mathcal{L}_{neg}$    & 93.61 & 95.30 & 95.16 &92.28 & 88.18 & 92.91 
\\ \hline
 \quad $\mathcal{L}_{w}+\mathcal{L}_{pos}$      & 93.03 & 95.18& 95.37 &92.60 & 86.87 & 92.61 
\\
 \quad $\mathcal{L}_{w}+\mathcal{L}_{neg}$      & 93.10 & 94.61 & \textbf{95.38}&92.42 & 88.65 & 92.83 
\\ \hline
\quad $\mathcal{L}_{SentiCSE}$         & \textbf{93.89}& \textbf{95.30} & 95.36 &\textbf{93.23}& \textbf{88.93}& \textbf{93.34}
\\ 
\hline
\end{tabular}}
\caption{Ablation of each component.} \label{table: without component}
\end{table}

\begin{table}[h!]
\centering \resizebox{0.48\textwidth}{!}{
\begin{tabular}{c|ccccc|c} \hline
\multicolumn{1}{l|}{$\lambda_w$} & \multicolumn{1}{c}{\textbf{IMDB}} & \multicolumn{1}{c}{\textbf{SST-2}} & \multicolumn{1}{c}{\textbf{Yelp-2}} & \multicolumn{1}{c}{\textbf{Amazon}}& \multicolumn{1}{c|}{\textbf{MR}} & \multicolumn{1}{c}{\textbf{Avg}} \\ \hline
0.15 & 93.89& \textbf{95.30}& 95.36 &93.42 & \textbf{88.93}& \textbf{93.88} \\ 
0.2  & 93.72 & 95.07 & \textbf{95.99}&93.01 & 87.90 & 93.14 \\
0.25 & \textbf{93.91}& 95.18 & 95.86 &\textbf{93.27}& 88.37 & 93.32\\ \hline
\end{tabular}} 
\caption{Impact of $\lambda_w$ on the performance.} 
\label{table:word-ratio}
\end{table}

\begin{table}[h!]
\centering \resizebox{0.48\textwidth}{!}{%
\begin{tabular}{c|ccccc|c} \hline
\multicolumn{1}{l|}{\textbf{Ratio}} & \multicolumn{1}{c}{\textbf{IMDB}} & \multicolumn{1}{c}{\textbf{SST-2}} & \multicolumn{1}{c}{\textbf{Yelp-2}}& \multicolumn{1}{c}{\textbf{Amazon}} & \multicolumn{1}{c|}{\textbf{MR}} & \multicolumn{1}{c}{\textbf{Avg}} \\ \hline
0.1 & 94.03& 95.18 & \textbf{95.86}& \textbf{93.69}& \textbf{89.49}& \textbf{93.65}\\
0.5 & 93.89 & \textbf{95.30}& 95.36 & 93.23 & 88.93 & 93.31 \\ 
0.9 & \textbf{94.06}& 94.84 & 95.46 & 93.19 & 88.37 & 93.18 \\ \hline
\end{tabular}} 
\caption{Sensitivity test on the masking ratio.}
\label{table:mask-ratio}
\end{table}

\begin{figure}[t]
    \centerline{\includegraphics[width=0.45\textwidth]{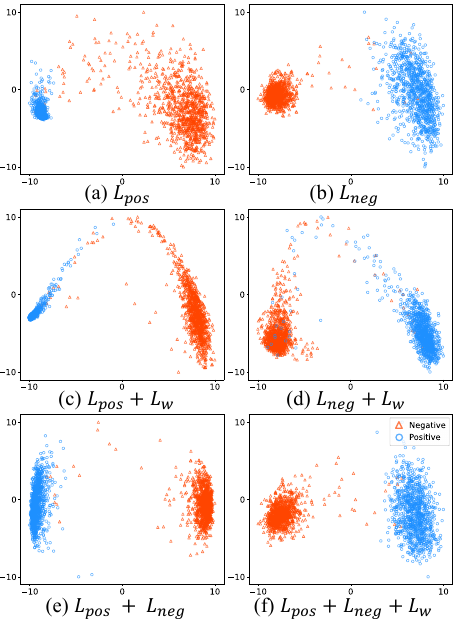}}
    \caption{Embedding visualizations of each ablation. PCA is used for visualization.}
   \label{fig:loss_ablation}
\end{figure}

First, we examined the impact of each component in SentiCSE. Table~\ref{table: without component} shows the performance of each ablation model on linear probing. In the sentence-level objective, we observed that using only $\mathcal{L}_{neg}$ provides a better result than using only $\mathcal{L}_{pos}$. This might be because of the phenomenon that diverse and rich vocabulary is used in negative reviews, while relatively consistent vocabulary is used in positive reviews~\cite{fang2015sentiment}. Using $\mathcal{L}_{pos}$ and $\mathcal{L}_{neg}$ together provides higher performance than using only one of them, which confirms the effectiveness of taking both polarity labels (positive and negative) as anchors. Adding $\mathcal{L}_{w}$ to each sentence-level objective consistently improves performance, which confirms that understanding sentence sentiment at word-level benefits learning sentiment representations. Furthermore, Figure~\ref{fig:loss_ablation} shows the visualization results obtained from each ablation. Notably, we can observe that contrastive learning using two sentences of different sentiment polarities as each anchor leads to higher quality representations compared to using only one sentence of a single sentiment polarity. We also confirm that the word-level loss plays an important role.

Next, we report the performance of SentiCSE according to varying $\lambda_w$ and the masking ratio used for the word-level objective. We tested different masking ratios with \{0.1, 0.5, 0.9\} and $\lambda_w$ with \{0.15, 0.2, 0.25\}. Tables~\ref{table: without component},~\ref{table:word-ratio} and~\ref{table:mask-ratio} report the linear probing performance in each case. We observed that their impact was not very significant.

Finally, we qualitatively evaluate the sentiment representation quality of SentiCSE through a case study. We fed one positive and one negative statement each about the taste of the food, the mood in the restaurant, and the friendliness of the staff for a particular restaurant into the models to obtain embeddings. We then searched for two sentences with the highest embedding similarity to the query, "The atmosphere of the restaurant is good", which is another positive statement about the atmosphere of the restaurant. The results, compared across the baselines, are shown in Table~\ref{table:sentences_example}. For SentiBERT and SentiWSP, the most similar sentence is positive, but the next similar sentence is a negative review about the service of the staff, which is contextually similar to the atmosphere of the restaurant. In the case of SentiLARE, the first sentence was a positive review, but the second was a negative review about the mood in the restaurant. SentiX recognized the negative review as the most similar sentence. On the other hand, our SentiCSE correctly identified positive reviews about the restaurant, highlighting its promising quality of sentence representation compared to the baselines in terms of sentiment analysis.

\begin{table}[t]
\centering
\resizebox{0.48\textwidth}{!}{%
\begin{tabular}{ll}
\Xhline{2\arrayrulewidth}
\multicolumn{1}{c}{\textbf{Query Sentence}}                   \\ \hline 
\multicolumn{1}{l}{The atmosphere of the restaurant is good.} \\ \hline \hline 
\multicolumn{1}{c}{\textbf{SentiBERT}}                        \\  \hline
1) The food is delicious.                                     \\
2) The service at the restaurant is discourteous.             \\\hline
\multicolumn{1}{c}{\textbf{SentiX}}                           \\ \hline
1) The service at the restaurant is discourteous.             \\
2) The food is delicious.  \\\hline
\multicolumn{1}{c}{\textbf{SentiLARE}}                        \\ \hline
1) The restaurant has attentive and friendly staff.                                        \\
2) The restaurant lacks a good ambiance.                \\\hline
\multicolumn{1}{c}{\textbf{SentiWSP}}                        \\ \hline
1) The food is delicious.                                        \\
2) The service at the restaurant is discourteous.                \\\hline
\multicolumn{1}{c}{\textbf{SentiCSE}}                         \\ \hline
1) The food is delicious.                                        \\
2) The restaurant has attentive and friendly staff.    \\ \Xhline{2\arrayrulewidth}
\end{tabular}%
}
\caption{From several positive and negative reviews about a restaurant, the two sentences with the highest embedding similarity to the query were identified by each model. Cosine similarity was used as the similarity metric.}
\label{table:sentences_example}
\end{table}

\section{Related Work}
This section focuses on PLMs specially designed for sentiment analysis because this line of research is the most relevant to our work.

SentiBERT~\citep{yin2020sentibert} embeds a two-level attention mechanism on top of the BERT representation, and introduces the binary constituency parse tree in order to capture phrase-level compositional sentence semantics.
SentiLARE (Sentiment-Aware Language Representation Learning)~\citep{ke2019sentilare} introduces sentiment-related linguistic knowledge obtained from SentiWordNet into PLMs, aiming at constructing knowledge-aware language representations. 
SKEP (Sentiment Knowledge Enhanced Pre-training)~\citep{tian-etal-2020-skep} designs three objective functions that predict sentiment-related knowledge to construct a unified sentiment representation for multiple downstream sentiment analysis tasks.
SentiX~\citep{zhou2020sentix} explores domain-invariant sentiment knowledge from large-scale review dataset and designs the cross-domain sentiment classification tasks.

Very recently, SimCSE~\citep{gao2021simcse} successfully introduces the contrastive learning paradigm to sentence embedding. Motivated by this work, sentence-level contrastive learning has been adopted to sentiment analysis research. 
SCAPT (Supervised ContrAstive Pre-Training)~\citep{li-etal-2021-learning-implicit} aligns the representation of implicit sentiment expressions with the same polarity in order to capture both implicit and explicit sentiment orientation. 
SentiWSP (Sentiment-Aware Word and Sentence Level Pre-training)~\citep{fan-etal-2022-sentiment} combines the word-level task that adopts MLM in sentiment word prediction and the sentence-level task which employs contrastive learning. 

We summarize the differences between the previous works and our method as follows: (1) In the word-level task, we focus on matching the polarity of words, rather than matching the words themselves. (2) In the sentence-level task, we employ \textit{supervised} contrastive learning that enables us to specify negatives to benefit from constructing sentiment representations. (3) We design the SgTS task to evaluate the quality of representations. 
\section{Conclusion}
This paper presents SgTS (Sentiment-guided Textual Similarity), a metric for evaluating sentiment representations. We then propose SentiCSE which focuses on the understanding sentiment of sentences via the word-level and sentence-level objectives, whose representation quality is guaranteed by SgTS. Experimental results confirm that SentiCSE shows higher than or comparable performance to the state-of-the-art baselines while requiring much less amount of data. We believe that our work sheds light on the importance of the quality of sentiment representations in the downstream sentiment analysis tasks with limited labeled data.

\section{Acknowledgement}
This work was partly supported by (1) the National Research Foundation of Korea (NRF) grant funded by the Korea government (*MSIT) (No.2018R1A5A7059549) and (2) the Institute of Information \& communications Technology Planning \& Evaluation (IITP) grant funded by the Korea government (MSIT) (No.2020-0-01373,Artificial Intelligence Graduate School Program (Hanyang University)). *Ministry of Science and ICT

\section{References}
\label{sec:reference}
\bibliographystyle{lrec-coling2024-natbib}
\bibliography{lrec-coling2024-example}

\clearpage
\appendix

\section{Appendix for SentiCSE}
\subsection{Hyper-parameter Settings}
Tables~\ref{tab:hyper-param-senticse} and~\ref{tab:hyper-param-fewshot} show the values of hyper-parameters used in the pre-training of SentiCSE and in the few-shot setting, respectively. For the linear probing setting, we borrowed the configurations of SentEval \citep{conneau-kiela-2018-senteval}. More details can be found in our code uploaded to the submission system.
\begin{table}[h!]
\centering{ 
\begin{tabular}{lc} \hline
\multicolumn{1}{l}{\textbf{Parameters}} & \multicolumn{1}{c}{\textbf{Values}} \\ \hline
Max sequence length & 128   \\ 
Batch size          & 64    \\ 
Learning rate       & 1e-5  \\ 
Number of steps    & 20,000 \\ 
Pooler type         & cls   \\
Temperature          & 0.05  \\
$\lambda_w$          & 0.15  \\
Hard negative weight     & 1     \\
Masking ratio   & 0.1   \\
\hline
\end{tabular}} 
\caption{Hyper-parameters for pre-training of SentiCSE.}
\label{tab:hyper-param-senticse}
\end{table}

\begin{table}[h!]
\centering{ 
\begin{tabular}{lc} \hline
\multicolumn{1}{l}{\textbf{Parameters}} & \multicolumn{1}{c}{\textbf{Values}} \\ \hline
Batch size          & 16    \\ 
Learning rate       & 1e-5  \\ 
Number of epochs    & 100 \\ 
Evaluation steps    & 20 \\ 
Weight decay        & 0.01 \\ 
\hline
\end{tabular}} 
\caption{Hyper-parameters for the few-shot setting.}
\label{tab:hyper-param-fewshot}
\end{table}

\subsection{More Analysis on SentiCSE}
In the main manuscript, we report the results obtained when MR data is adopted in SentiCSE as a source domain. Here, we tested other data as the source domain of SentiCSE. Table~\ref{table:dataset_ablation} reports the performance obtained from each case on the linear probing, respectively. We observed that the average performance was higher in the order when the source domain was MR, SST-2, Yelp-2, and IMDB. This order is very interesting because, as shown in Table 1 in the main manuscript, it has the same order with the `\% of SentiWords' of each dataset. This implies the importance of the amount of sentiment information contained in each sentence when choosing a source domain for pre-training.

\begin{table}[h!]
\centering \resizebox{0.48\textwidth}{!}{%
\begin{tabular}{l|cccc|c} \hline
{Source$\setminus$Target} & \multicolumn{1}{l}{\textbf{IMDB}} & \multicolumn{1}{l}{\textbf{SST-2}} & \multicolumn{1}{l}{\textbf{Yelp-2}} & \multicolumn{1}{l|}{\textbf{MR}} & \multicolumn{1}{l}{\textbf{Avg.}} \\ \hline
\textbf{IMDB}    &   $94.21^*$  &   89.22   &   95.79   & 86.49     & 91.43 \\
\textbf{SST-2}   &   90.62    &   $94.72^*$ &   94.86   & \textbf{89.87}     & 92.52 \\
\textbf{Yelp-2}  &   93.67    &   87.50   &   $97.93^*$ & 86.59     & 91.42 \\
\textbf{MR}      &   \textbf{94.28}    &   \textbf{95.30}   &   \textbf{96.27}   & $89.02^*$     & \textbf{93.72} \\  \hline
\end{tabular}}
\caption{Linear probing performance of SentiCSE on each source-target data combination. For each data column, the accuracy with * indicates the result from a model that was pre-trained on the same dataset as its source domain.}
\label{table:dataset_ablation}
\end{table}

Furthermore, we also implemented BERT-based SentiCSE and compared it with the BERT-based baselines: SentiBERT and SentiX. We also compared BERT itself and BERT-based SimCSE with ours. Overall, a similar trend is observed when we compared the RoBERTa-based models including SentiLARE.

Next, we examined the impact of $\alpha$ used in the sentence-level objective on the performance. Table~\ref{table:ablation_alpha} shows the linear probing results with regards to different $\alpha$ values.
\begin{table}[ht!]
\centering \resizebox{0.44\textwidth}{!}{%
\begin{tabular}{c|cccc|c} \hline
\multicolumn{1}{l|}{\textbf{$\alpha$}} & \multicolumn{1}{c}{\textbf{IMDB}} & \multicolumn{1}{c}{\textbf{SST-2}} & \multicolumn{1}{c}{\textbf{Yelp-2}} & \multicolumn{1}{c|}{\textbf{MR}} & \multicolumn{1}{c}{\textbf{Avg.}} \\ \hline
0 &  94.02 & \textbf{95.30} & 96.30 & \textbf{89.12} & 93.69 \\
1 &  \textbf{94.06} & 94.95 & \textbf{96.27} & 88.93 & 93.55 \\
2 &  93.99 & 95.07 & \textbf{96.27} & 88.18 & 93.38 \\ \hline
\end{tabular}} 
\caption{Performance depending on different $\alpha$.}
\label{table:ablation_alpha}
\end{table}

\begin{table*}[ht!]
\centering
\resizebox{\textwidth}{!}{%
\begin{tabular}{ccccccccc}
\Xhline{3\arrayrulewidth}
\multicolumn{1}{c}{Loss} & Dataset & RoBERTa & SimCSE & SentiBERT & SentiX & SentiLARE & SentiWSP & SentiCSE \\
\hline
\multirow{4}{*}{Alignment} & IMDB & 0.01 & 0.24 & 0.32 & 1.12 & 0.21 & 0.23 & 1.37 \\
 & SST-2 & 0.00 & 0.46 & 0.22 & 1.13 & 0.09 & 0.32 & 1.65 \\
 & Yelp-2 & 0.01 & 0.38 & 0.30 & 1.23 & 0.43 & 0.23 & 1.34 \\
 & MR & 0.00 & 0.45 & 0.23 & 1.11 & 0.09 & 0.30 & 1.61 \\
\hline
\multirow{4}{*}{Uniformity} & IMDB & -0.01 & -0.48 & -0.59 & -1.94 & -0.38 & -0.43 & -1.07 \\
 & SST-2 & -0.01 & -0.91 & -0.41 & -2.01 & -0.19 & -0.60 & -1.02 \\
 & Yelp-2 & -0.01 & -0.72 & -0.56 & -1.89 & -0.71 & -0.43 & -1.06 \\
 & MR & -0.01 & -0.89 & -0.44 & -2.00 & -0.18 & -0.58 & -1.19 \\
 \Xhline{3\arrayrulewidth}
\end{tabular}%
}
\caption{Representation quality measured by Alignment and Uniformity. The lower the scores, the better the quality.}
\label{tab:align_and_uniform}
\end{table*}

\subsection{More Analysis on SgTS}
\begin{figure}[ht!]
    \centerline{\includegraphics[width=0.5\textwidth]{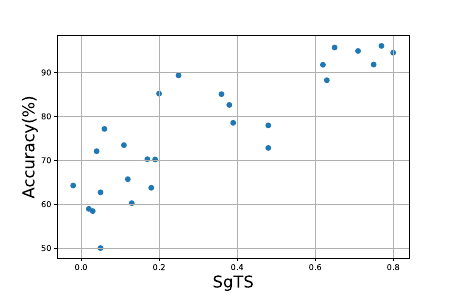}}
    \caption{Correlation between the SgTS score and the few-shot accuracy.}
   \label{fig:sgts_corr}
\end{figure}

As mentioned in the main manuscript, we observed the high correlation between our SgTS metric and the few-shot performance. We plot the SgTS scores and corresponding few-shot accuracy as shown in Figure~\ref{fig:sgts_corr} (the Spearman correlation $\rho=0.96$).

\begin{figure}[ht!]
    \centerline{\includegraphics[width=0.48\textwidth]{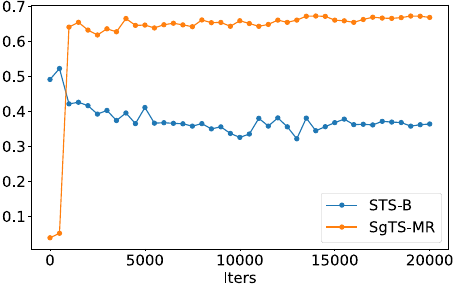}}
    \caption{Behavior of the SgTS and STS scores in pre-training.}
   \label{fig:sgTS_stsb}
\end{figure}

Next, we examined the behavior of the SgTS scores compared to the conventional STS as the pre-training of SentiCSE progresses. Figure~\ref{fig:sgTS_stsb} shows the results. Each score was obtained at every 500 step. We observed that the SgTS score reached to a high value in the very initial stage, which indicates the effectiveness of our pre-training objectives. We also observed that in the first half steps, SgTS gradually increased while STS gradually decreased. This implies the difference between sentiment-favorable representation and semantic representation. However, after convergence of the two scores, we observed that the STS score converges around 0.4, which demonstrates that understanding semantic information is still helpful in constructing high-quality sentiment representations.

\subsection{Alignment and Uniformity Metrics}
In terms of evaluation methods for representation quality, Alignment and Uniformity have been known as key properties in contrastive learning \citep{wang2020understanding}. The two factors take alignment between the pairs in the same class and uniformity of the representation space \citep{gao2021simcse}, respectively. They can measure the quality of learned embeddings. Table~\ref{tab:align_and_uniform} shows their scores measured on the constructed representations obtained from each model. It was difficult to find high correlation between the quality of sentiment representations visualized via PCA and the Alignment and Uniformity scores. In our future work, we plan to develop a novel evaluation metric for sentiment representations based on Alignment and Uniformity. 

\end{document}